\documentclass[runningheads]{llncs}
\usepackage[T1]{fontenc}
\usepackage{graphicx}
\usepackage{subcaption}
\usepackage{amsmath} 
\usepackage{booktabs}
\usepackage{xcolor}
\usepackage{amsfonts}
\usepackage{mathrsfs}
\usepackage{bbm}

\begin{document}
\title{Efficient Multi-Task Inferencing: Model Merging with Gromov-Wasserstein Feature Alignment
% a Shared Backbone and Lightweight Task-Specific Adapters for Automatic Scoring
}

\titlerunning{Model Merging with GW Alignment}

\author{ Luyang Fang\inst{1,3} \and
Ehsan Latif\inst{1,2} \and
Haoran Lu\inst{3}
Yifan Zhuo\inst{1,4} \and
Ping Ma\inst{3}\textsuperscript{,*} \and
Xiaoming Zhai\inst{1,2}\textsuperscript{,*}
}
\institute{AI4STEM Education Center, Athens, GA, USA \and
Department of Mathematics, Science, and Social Studies Education, University of Georgia, Athens, GA, USA \and
Department of Statistics, University of Georgia, Athens, GA, USA \and
School of Computing, University of Georgia, Athens, GA, USA \\
*\email{\{pingma,xiaoming.zhai\}@uga.edu}}

\maketitle              % typeset the header of the contribution

\begin{abstract}
Automatic scoring of student responses enhances efficiency in education, but deploying a separate neural network for each task increases storage demands, maintenance efforts, and redundant computations. To address these challenges, this paper introduces the Gromov-Wasserstein Scoring Model Merging (GW-SMM) method, which merges models based on feature distribution similarities measured via the Gromov-Wasserstein distance. Our approach begins by extracting features from student responses using individual models, capturing both item-specific context and unique learned representations. The Gromov-Wasserstein distance then quantifies the similarity between these feature distributions, identifying the most compatible models for merging. Models exhibiting the smallest pairwise distances, typically in pairs or trios, are merged by combining only the shared layers preceding the classification head. This strategy results in a unified feature extractor while preserving separate classification heads for item-specific scoring. 
We validated our approach against human expert knowledge and a GPT-o1-based merging method. GW-SMM consistently outperformed both, achieving higher micro F1 score, macro F1 score, exact match accuracy, and per-label accuracy. The improvements in micro F1 and per-label accuracy were statistically significant compared to GPT-o1-based merging ($p=0.04,\ p=0.01$). Additionally, GW-SMM reduced storage requirements by half without compromising much accuracy, demonstrating its computational efficiency alongside reliable scoring performance.

\keywords{Multi-task learning \and Efficient inference \and Automatic Scoring \and Deep Learning \and Optimal Transport}
\end{abstract}

\section{Introduction}

With rapid advancements in deep neural networks \cite{achiam2023gpt,brown2020language,devlin2018bert}, deep learning-based automated scoring systems offer a powerful means to efficiently assess students' complex thinking and knowledge-in-use, provide timely feedback, and reduce instructor workload \cite{latif2024fine,lee2024applying}.
Traditionally, separate neural networks are trained for each task in an assessment, but deploying multiple models in resource-constrained online learning environments is difficult due to high storage and computational demands \cite{zhai2022applying}, as well as the ongoing need for updates and maintenance.  Moreover, when tasks share similar concepts or structures, training separate models fails to exploit common patterns in student responses, limiting efficiency and generalization.

Several strategies have been proposed to address the aforementioned challenges, including multi-task learning \cite{geden2020predictive,zhang2021survey}, parameter-efficient fine-tuning \cite{ding2023parameter,hu2021lora,katuka2024investigating,latif2024efficient}, and knowledge distillation \cite{fang2024bayesian,gou2021knowledge}. However, they may still be costly or impractical when pre-trained models already exist for individual tasks. In such cases, model merging, which consolidates multiple fine-tuned models into one network, reduces storage and computation while preserving performance. Unlike fine-tuning, the merged model directly integrates knowledge from each individual model. This method is particularly advantageous when multiple pre-trained models exist, but a more efficient, unified solution is needed.

However, a key challenge arises when model merging scales beyond three models, often leading to impractical outcomes and performance issues \cite{yamada2024toward}. To address this, we propose a novel strategy that quantifies similarity among pre-trained models using the Gromov-Wasserstein (GW) distance, guiding which models should be merged. 
Our Gromov-Wasserstein Scoring Model Merging (GW-SMM) framework first extracts features from student responses of each model to capture both question-level context and model-specific representations. Unlike methods that compare only model parameters or rely on surface-level question similarities, our feature-based approach effectively captures the interplay between item content and learned model nuances.
We then apply the GW distance, an optimal transport-based metric that aligns distributions while preserving structural relationships \cite{memoli2014gromov,peyre2019computational,zhang2023projection}. One key advantage of the GW distance over the standard Wasserstein metric is its ability to compare distributions even when they reside in different latent spaces \cite{peyre2016gromov}. This allows us to accommodate feature embeddings that may be learned in structurally distinct but meaningful ways across models, ensuring a more principled merging strategy.

The proposed method supports resource-efficient AI for education and other cost-sensitive domains, contributing to advancements in scalable, efficient model deployment. Our key contributions are summarized below:
\begin{enumerate}
    \item \textbf{Efficient Model Merging via GW Distance:} We develop a novel model merging method using GW distance to align feature distributions across models. This approach ensures structural consistency between feature spaces while significantly reducing inference costs compared to traditional task-specific models.
    \item \textbf{Effective Feature Extraction:} We extract features by combining representations from pre-trained individual models, capturing both question-level context and model-specific learning patterns. This feature-based alignment offers a more robust merging strategy than parameter-based approaches.
    \item \textbf{Performance Evaluation:} We evaluate GW-SMM across automatic scoring tasks, demonstrating its effectiveness in preserving predictive accuracy while reducing computational overhead. 
\end{enumerate}

%-------------------------------------
\section{Methodology}

\begin{figure}[t]
    \centering
    \includegraphics[width=1\linewidth]{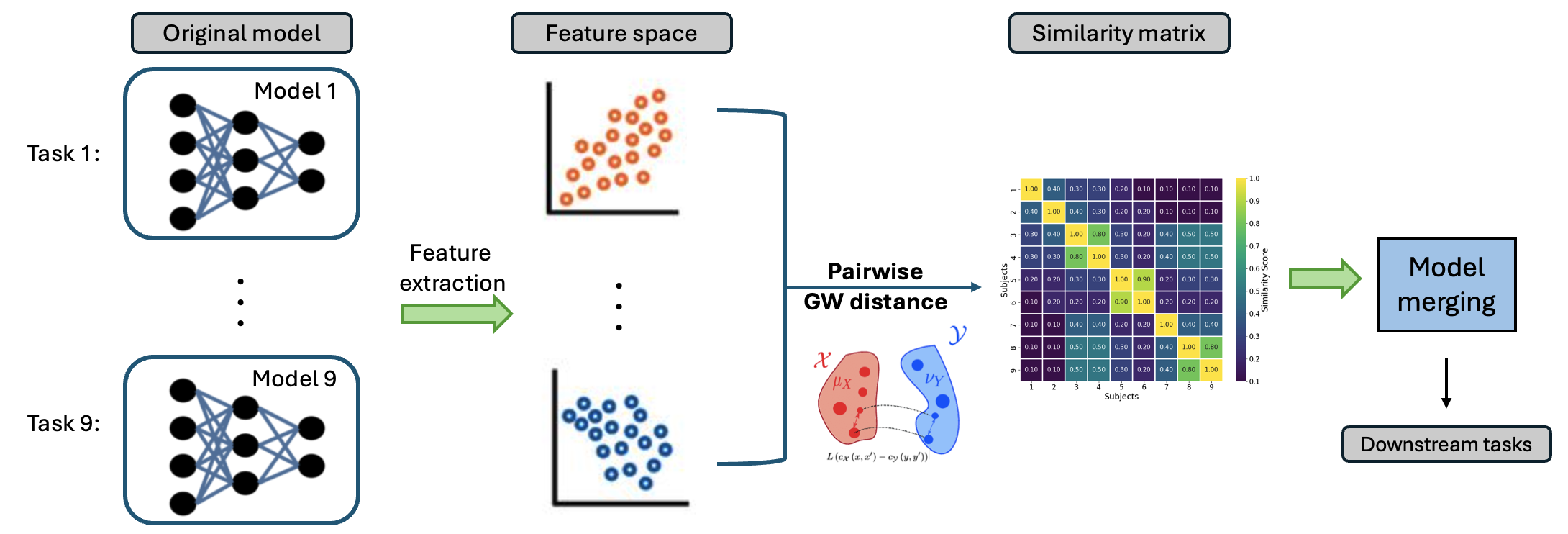}
    \caption{Workflow of the developed GW-SMM method.}
    \label{fig:workflow}
    % \vspace{-11pt}
\end{figure}

This section first introduces the GW distance, which is used to determine task similarity. Next, we describe the selection of the merging plan based on the similarity matrix, followed by an explanation of the merging algorithm applied. Workflow of GW-SMM is available in Figure \ref{fig:workflow}.

\subsection{Gromov-Wasserstein Distance}
We use the GW distance to compare feature distributions of student responses across items, even when these features lie in distinct metric spaces. Formally, let $\mu$ and $\nu$ denote probability measures over feature spaces $(\mathcal{X}, c_{\mathcal{X}})$ and $(\mathcal{Y}, c_{\mathcal{Y}})$, respectively. Here, features may come from heterogeneous pre-trained models encoding task-level context and model-specific learning patterns.

A central concept in GW distance is the coupling $\pi \in \Pi(\mu, \nu)$, which is a joint distributions preserving the marginals $\mu$ and $\nu$ \cite{villani2021topics}:
\begin{small}
    \begin{equation}
    \Pi(\mu, \nu)= \left\{\pi \in \mathscr{P}\left(\mathcal{X} \times \mathcal{Y}\right): 
    \pi\left(A \times \mathcal{Y}\right)=\mu(A), \forall A \subset \mathcal{X}, \pi\left( \mathcal{X}\times B \right)=\nu(B), \forall B \subset \mathcal{Y} \right\} .
\end{equation}
\end{small}
% \begin{align}
%     \Pi(\mu, \nu) = \bigg\{ \pi \in \mathscr{P}(\mathcal{X} \times \mathcal{Y}): \quad 
%     & \pi(A \times \mathcal{Y}) = \mu(A), \quad \forall A \subset \mathcal{X}, \notag \\ 
%     & \pi(\mathcal{X} \times B) = \nu(B), \quad \forall B \subset \mathcal{Y} 
%     \bigg\}.
% \end{align}
A coupling $\pi \in \Pi(\mu, \nu)$ is a joint distribution of $\mu$ and $\nu$ such that two particular marginal distributions of $\pi$ are equal to $\mu$ and $\nu$, separately.
The standard Wasserstein distance between two feature distributions can be formulated as finding the optimal coupling that minimizes the expected cost under a suitable cost function $c$,
\begin{equation}\label{eq:W}
    \pi^* = \min _{\pi \in \Pi(\mu, \nu)} L(\pi) = \min _{\pi \in \Pi(\mu, \nu)} \int_{\mathcal{X} \times \mathcal{Y}} c(x,y) \mathrm{~d} \pi(x, y),
\end{equation}
where $c$ is a positive lower semi-continuous cost function $c: \mathcal{X} \times \mathcal{Y} \rightarrow \mathbb{R}^{+}$. However, defining such a cost function $c$ becomes difficult when $\mathcal{X}$ and $\mathcal{X}$ are non-aligned or differ in dimensionality and semantics.

To address this limitation, we employ the GW distance, which focuses on structural discrepancies within each distribution rather than requiring a direct cost function across spaces. Given two measurable functions $c_{\mathcal{X}}: \mathcal{X} \times \mathcal{X} \rightarrow \mathbb{R}$ and $c_\mathcal{Y}: \mathcal{Y} \times \mathcal{Y} \rightarrow \mathbb{R}$, the GW distance is defined as:
\begin{equation}\label{eq:GW}
GW(c_\mathcal{X}, c_\mathcal{Y}, \mu, \nu):=\inf _{\pi \in \Pi(\mu, \nu)} \int_{\mathcal{X}^2 \times \mathcal{Y}^2} L\left(c_\mathcal{X}\left(x, x^{\prime}\right)-c_\mathcal{Y}\left(y, y^{\prime}\right)\right) \mathrm{~d} \pi(x, y) \mathrm{d} \pi\left(x^{\prime}, y^{\prime}\right) ,
\end{equation}
where $L$ is a distance measure with common choices $L(a,b)=|a-b|^p$, $p\geq 1$.

Since this optimization problem is non-convex and non-smooth, we solve it using iterative methods with entropic regularization \cite{peyre2016gromov}. At the $r^{th}$ iteration, we update the coupling $\pi$ by solving:
\begin{equation}
\pi^{(r+1)}:=\arg \min _{\pi \in \Pi(\mu, \nu)} \left[ GW^{(r)} + \epsilon \mathcal{H}(\pi) \right] ,
\end{equation}
where $\mathcal{H}(\pi)$ is the Shannon entropy of $\pi$, and $\epsilon>0$ controls the regularization strength.
% where $\mathcal{H}(\pi)=-\int_{\mathcal{X} \times \mathcal{Y}} \pi(x, y) \log \pi(x, y) \mathrm{d} x \mathrm{~d} y$ denotes the Shannon entropy of the coupling $\pi$, and $\varepsilon>0$ controls the regularization strength.
The term 
\begin{equation}
    GW^{(r)}:= \int_{\mathcal{X}^2 \times \mathcal{Y}^2} L\left(c_\mathcal{X}\left(x, x^{\prime}\right)-c_\mathcal{Y}\left(y, y^{\prime}\right)\right) \mathrm{~d} \pi^{(r)}(x, y) \mathrm{d} \pi^{(r)}\left(x^{\prime}, y^{\prime}\right)
\end{equation}
quantifies structural mismatches in feature-space geometry based on the previous iteration’s coupling. By incorporating entropy, the algorithm attains a smooth and numerically stable solution through Sinkhorn iterations \cite{cuturi2013sinkhorn,sinkhorn1967concerning}, enabling efficient approximation of the GW distance in our similarity comparison setting.

%-----
\textbf{Feature Extraction. } 
Given $T$ different tasks (in our case, exam questions), we have $T$ fine-tuned models that capture the unique response patterns of different assessment tasks.
We extract features for each model using its corresponding dataset. For each item, we pass its data through the fine-tuned model and collect the activations from a selected intermediate layer. A common choice in transformer-based architectures, such as BERT, is to use the final hidden layer before the classification head, as it retains high-level semantic representations of the input while avoiding distortions introduced by the final task-specific decision layer. Prior research has demonstrated that leveraging these deep representations provides a more robust measure of model similarity compared to directly comparing model parameters or lower-layer representations \cite{devlin2018bert,merchant2020happens}.

% \textbf{Threshold to determine which models should be merged: (describe here or move to the experiments part)} 

\subsection{Determination of the Merging Plan } \label{sec:mergingplan}
To determine which models should be merged, we construct a pairwise similarity matrix based on the feature distributions extracted by each model. We first compute the GW distance between every pair of models to quantify the structural alignment between their feature spaces. Since the GW distance measures dissimilarity, we transform it into a similarity score $ S_{i,j} $ using the formula, 
$
S_{i,j} = 1 - \frac{d_{i,j} - d_{\min}}{d_{\max} - d_{\min}},
$
where $ d_{i,j} $ is the GW distance between models $ i $ and $ j $, and $ d_{\min} $ and $ d_{\max} $ are the minimum and maximum GW distances in the computed matrix. This transformation rescales the distances into similarity scores ranging from 0 (least similar) to 1 (most similar), allowing for a more interpretable and structured merging decision process.

Given the similarity matrix and the number $T'$ of final merged models, we propose to determine the optimal merging plan by formulating an objective function that balances consolidation efficiency and task-specific accuracy preservation. We find the merging plan that maximizes the loss function,
$L(\mathcal{M}) = \sum_{(i, j) \in \mathcal{M}} S_{i,j}$,
where $ \mathcal{M} $ is the set of merged pairs or groups, $ S_{i,j} $ represents the similarity score between models $ i $ and $ j $.
To optimize this objective, we search for the merging plan $ \mathcal{M}^* $ that minimizes $ L(\mathcal{M}) $ over all possible merge plans that will result in $T'$ final merged models.

\subsection{Model Merging}

In our approach, we combine several fine-tuned models into a single unified model to reduce storage and deployment costs while maintaining high task performance. Although basic methods such as parameter averaging (e.g., \cite{ilharco2023editing}), Fisher-weighted merging \cite{matena2021merging}, and task arithmetic \cite{ilharco2023editing} have been explored, our work focuses on the \texttt{TIES-MERGING} framework \cite{yadav2023ties} that particular fits our multi-task scenario.

Rather than simply averaging parameters or directly combining task-specific updates, \texttt{TIES-MERGING} enhances the merging process by explicitly aligning model representations and pruning redundant or conflicting parameters. Let $\theta_t$ denote the parameters of the fine-tuned model for task $t$, and let $\theta_0$ represent the shared backbone. We first compute the task-specific update as $\tau_t = \theta_t - \theta_0.$

Instead of merging these updates directly, we align the feature spaces of individual models using techniques such as optimal transport. This alignment ensures that similar features across models are brought into correspondence, leading to a more coherent integration of the learned representations.

After alignment, a pruning mechanism is applied to eliminate redundant or conflicting parameters. This step stabilizes the merged model by preserving only the essential task-specific information and mitigating destructive interference. The final merged parameters are obtained by
\begin{equation}
\theta_{\text{merged}} = \theta_0 + \sum_{t=1}^{T} \lambda_t \tau_t,
\end{equation}
where the coefficients $\lambda_t$ are determined by the alignment and pruning process.

By combining representation alignment with targeted pruning, \texttt{TIES-MERGING} effectively leverages shared knowledge across tasks while maintaining nuanced task-specific distinctions. This makes our method particularly well-suited for applications such as automated scoring systems, where both accuracy and efficiency are critical.

%================================
\section{Dataset Details}\label{sec:dataset}

This research utilizes pre-existing datasets, incorporating responses from middle school students that have been evaluated by experts for nine multi-label assessment tasks from the PASTA project \cite{Harris2024Creating,PASTA2023}. These assessment tasks are specifically crafted to evaluate middle school students' ability to apply multi-label knowledge in explaining scientific phenomena. The NGSS framework aims to facilitate students in developing applied knowledge across educational levels by integrating disciplinary core ideas (DCIs), crosscutting concepts (CCCs), and science and engineering practices (SEPs) within K-12 performance expectations. The assessment tasks in this study align with the NGSS middle school-level expectations: students must analyze and interpret data to determine whether substances possess identical properties \cite{national2013next}. This expectation requires students to employ knowledge of the structure and properties of matter, chemical reactions (DCIs), and patterns (CCC) to effectively analyze and interpret data (SEP).

\begin{figure}[hbt!]
    \centering
    \includegraphics[width=0.99\linewidth]{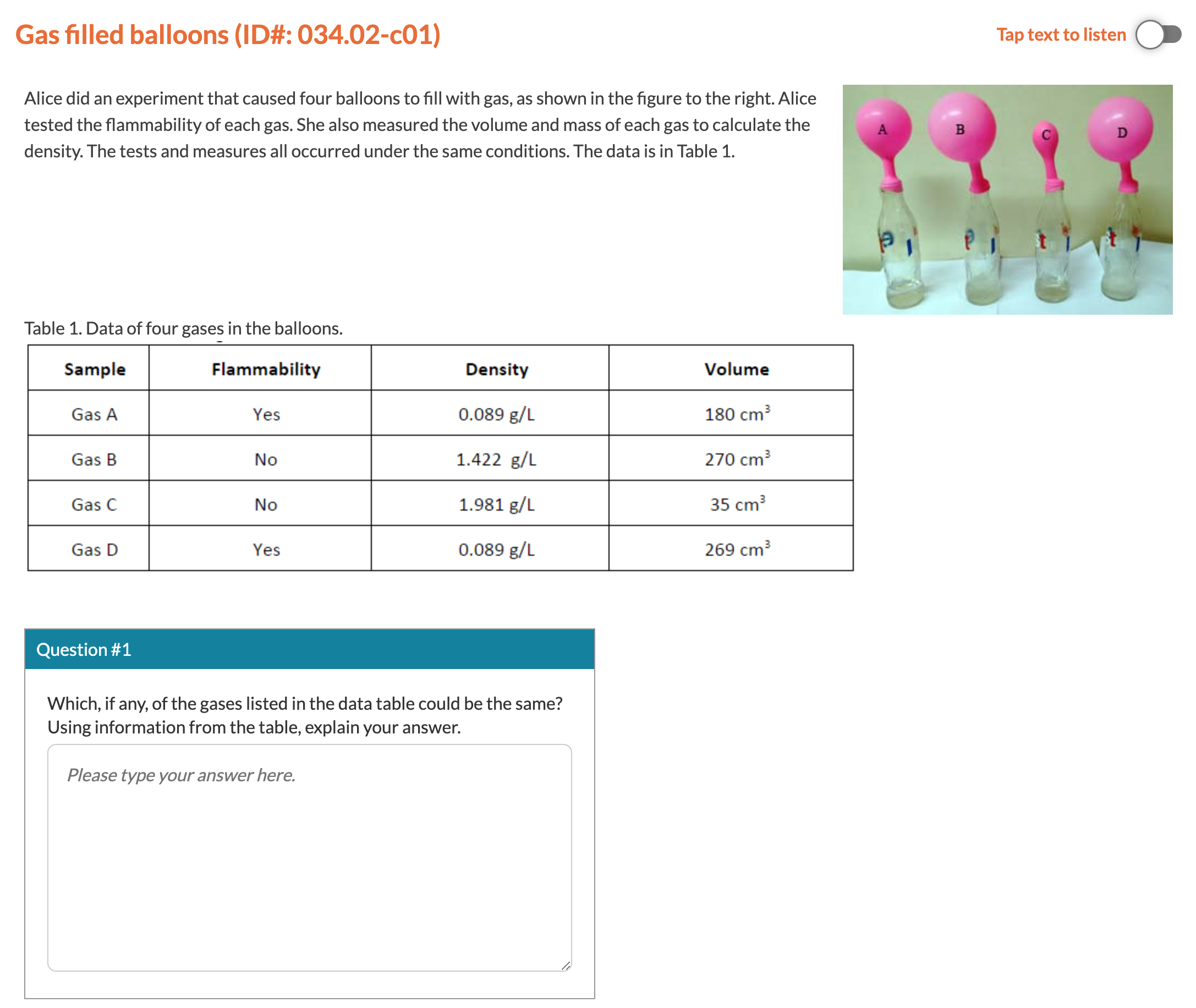}
    \caption{Illustrative Multi-label Task: Gas-Filled Balloons}
    \vspace{-5pt}
    \label{fig:gas_filled_ballons}
\end{figure}

A total of 1,200 students in grades 6–8 participated in this study. Middle school teachers across the U.S. invited their students to engage with open-ended NGSS-aligned science tasks \cite{zhai2022applying}. After data cleaning, fewer than 1,200 responses remained per task (exact counts in Table~\ref{table:data}). Responses were randomly selected to form training, validation, and test sets for machine learning models. For privacy, all data was anonymized, and demographic details were unavailable. Nonetheless, due to the geographical diversity of participating teachers, the dataset is considered representative of the broader US middle school student population.

The assessment tasks were sourced from the Next Generation Science Assessment \cite{Harris2024Creating} and required students to apply fundamental \textit{chemistry} principles to real-world contexts. Falling under the physical sciences domain, specifically ``Matter and its Characteristics'', these tasks assess students' ability to analyze data and differentiate substances by their attributes. These tasks were designed to assess students' multi-dimensional thinking and provide educators with insights that could inform instructional strategies. Automated reports derived from rubric-based scoring highlight topics where students may require additional support.
For instance, in one task, students were asked to identify gases in an experiment by comparing their properties to those documented in a data table (refer to Fig.~\ref{fig:gas_filled_ballons}).  Successfully completing this task required understanding the structure and properties of matter, chemical reactions, and the ability to plan investigations while recognizing patterns in the data.

A structured scoring rubric was developed to encompass five response dimensions, corresponding to the science learning framework: SEP+DCI, SEP+CCC, SEP+CCC, DCI, and DCI. This rubric was designed to capture multi-dimensional cognitive processes \cite{He2024G}. Table~\ref{table:rubric_gas_filled_ballon} outlines the specific criteria for each category. Students were assessed simultaneously across multiple perspectives, receiving scores that reflected their understanding of DCIs, CCCs, and SEPs as aligned with the rubric. To enhance the validity of these multi-perspective rubrics, the research team collaborated with experienced science educators.

\begin{table}[hbt!]
\centering
\caption{Scoring rubric for task: Gas-filled balloons (Task 5).}
\resizebox{0.9\textwidth}{!}{ 
\begin{tabular}{l| l| p{9.5cm} }
\toprule
\textbf{ID} & \textbf{Perspective} & \textbf{Description} \\
\midrule
E1 & SEP+DCI &  Student states that Gas A and D could be the same substance. \\
% \midrule
E2 & SEP+CCC & Student describes the pattern (comparing data in different columns) in the table flammability data of Gas A and Gas D as the same. \\
% \midrule
E3 & SEP+CCC & Student describes the pattern (comparing data in different columns) in density data of Gas A and Gas D, which is the same in the table. \\
% \midrule
E4 & DCI & Student indicate flammability is one characteristic of identifying substances. \\
% \midrule
E5 & DCI & Student indicate density is one characteristic of identifying substances. \\
\bottomrule
\end{tabular}
}
\label{table:rubric_gas_filled_ballon}
\end{table}

\begin{table}[hbt!]
    \centering
    \caption{Dataset information for both multi-label and multi-class tasks}
    \resizebox{0.9\textwidth}{!}{ 
    \begin{tabular}{l | l| c| c| c}
        \toprule
         \textbf{ID}& \textbf{Item}& \textbf{No. Labels} & \textbf{Training size} & \textbf{Testing size} \\
        \midrule
        Task 1 & Anna vs Carla & 4 & 955 & 239 \\
        Task 2 & Breaking Down Hydrogen Peroxide & 4 & 666 & 167 \\
        Task 3 & Carlos Javier Atomic Model & 5 & 956 & 240 \\
        Task 4 & Dry Ice Model & 3 & 1111 & 278 \\
        Task 5 & Gas Filled Balloon & 3 & 958 & 240 \\
        Task 6 & Layers in Test Tube & 10 & 956 & 240 \\
        Task 7 & Model For Making Water & 5 & 836 & 210 \\
        Task 8 & Nami Careful Experiment & 6 & 653 & 164 \\
        Task 9 & Natural Sugar & 5 & 956 & 239 \\
        \bottomrule
    \end{tabular}
    }
    \label{table:data}
\end{table}

\section{Experimentation}

This section outlines the implementation details, experimental setup, and evaluation metrics employed to validate the proposed framework.

\subsection{Experimental Setup}

To automatically assess students' open-ended responses, we have first employed a multi-label binary classification approach using a fine-tuned BERT-based model \cite{devlin2018bert}. Given the complexity of students' explanations, which often span multiple dimensions of scientific understanding, a multi-label classification setup was necessary. Each response was independently evaluated across multiple rubric dimensions, resulting in multiple binary labels per response (as mentioned in Sec. ~\ref{sec:dataset}). We utilized the pre-trained BERT-base model from Hugging Face \cite{wolf2020transformers}, fine-tuned on our dataset using a task-specific classification head. The input to the model consisted of tokenized student responses, processed using the WordPiece tokenizer \cite{wu2016google}. Each response was mapped to a fixed-length sequence with special tokens ([CLS] and [SEP]) and subsequently passed through the BERT encoder.

For fine-tuning, we adopted a binary cross-entropy loss function applied independently to each label. The model was trained using the AdamW optimizer \cite{loshchilov2017decoupled} with a learning rate of $2e^{-5}$ and batch size of 32. To mitigate overfitting, dropout regularization \cite{srivastava2014dropout} was employed within the classification head, and early stop was employed based on the validation loss. The training process was conducted for some varying numbers of epochs (ranging from 10 to 20) on an NVIDIA GPU, leveraging early stopping based on validation loss.

% \lf{Haoran, please provide details about the extraction of features for similarity calculation}

For each fine-tuned model, we remove the classification head and run a forward pass on all relevant student responses for the corresponding item. We enable the output of hidden states and collect the activations from the final Transformer layer. Specifically, we extract the [CLS] token embedding of each sample, which retains high-level semantic information needed for subsequent clustering and merging. This procedure yields a matrix of embeddings of shape $(n_i, d)$, where $n_i$ is the total number of student responses for the item, and $d=768$ is the dimensionality of the final hidden representation. Further, we compute pairwise Euclidean distances among all embeddings and use the resulting distance matrix for model-merging decisions.

\subsection{Evaluation Metrics}

To assess the performance of different methods, we employ four evaluation metrics: Micro F1 Score, Macro F1 Score, Exact Match Accuracy, and Per-label Accuracy. These metrics provide a comprehensive assessment of multi-class classification performance, capturing overall predictive power and per-class balance. 
For all four metrics, higher values indicate better performance. 

\noindent\textbf{Micro F1 Score:}
The Micro F1 Score is calculated by aggregating the counts of true positives, false positives, and false negatives across all classes before computing the F1 score. It is defined as:
\begin{small}
    \begin{equation}
    \text{Micro F1} = \frac{2 \sum_{c=1}^C TP_c}{2 \sum_{c=1}^C TP_c + \sum_{c=1}^C FP_c + \sum_{c=1}^C FN_c} ,
\end{equation}
\end{small}
where $C$ is the total number of classes, and $TP_c$, $FP_c$, and $FN_c$ denote the true positives, false positives, and false negatives for class $c$. This metric assigns equal weight to each instance, making it suitable for imbalanced datasets.

\noindent\textbf{Macro F1 Score:}
The Macro F1 Score computes the F1 score separately for each class and then takes the unweighted average across all classes:
\begin{small}
    \begin{equation}
    \text{Macro F1} = \frac{1}{C} \sum_{c=1}^{C} \frac{2 TP_c}{2 TP_c + FP_c + FN_c} ,
\end{equation}
\end{small}
Unlike the Micro F1 Score, this metric gives equal importance to all classes, making it more sensitive to performance in underrepresented classes.

\noindent\textbf{Exact Match Accuracy:}
Exact Match Accuracy measures the proportion of instances for which the predicted label set exactly matches the ground truth label set:
\begin{small}
    \begin{equation}
    \text{Exact Match Accuracy} = \frac{1}{N} \sum_{i=1}^{N} \mathbbm{1}(\hat{Y}_i = Y_i) ,
\end{equation}
\end{small}
where $N$ is the total number of responses, $\hat{Y}_i$ is the predicted label set for instance $i$, and $Y_i$ is the ground truth label set. The indicator function $\mathbbm{1}(\cdot)$ returns 1 if the predicted and ground truth labels match exactly and 0 otherwise.

\noindent\textbf{Per-label Accuracy:}
Per-label Accuracy calculates the accuracy for each label independently and then averages across all labels:
\begin{small}
    \begin{equation}
    \text{Per-label Accuracy} = \frac{1}{C} \sum_{c=1}^{C} \frac{TP_c + TN_c}{TP_c + TN_c + FP_c + FN_c},
\end{equation}
\end{small}
where $TN_c$ represents the number of true negatives for class $c$. Unlike Exact Match Accuracy, this metric allows partial correctness to be recognized, making it more forgiving in multi-label classification settings.

%=========================
\section{Results}\label{sec:results}

To evaluate the effectiveness of the merging method, we assess its performance across nine tasks. For tasks where models are merged, performance is evaluated using the merged model. If a task's model remains unmerged, its performance is assessed using the original models. As a comparison, we evaluate GW-SMM against two methods to find similarities:
\begin{itemize}
    \item \textbf{Human Knowledge:} the merging plan is determined based on the task similarities identified by a human expert according to the task exploration.
    \item \textbf{GPT-o1:} the merging plan is generated by prompting GPT-o1 with task descriptions to identify task similarities.
\end{itemize}

We compare the average performance of the three merging methods across the four evaluation criteria and also assess their performance relative to the original (pre-merge) models. Additionally, we conduct paired t-tests to statistically compare the effectiveness of different methods.

We note that Human GW-SMM is data-driven, leveraging all student responses to infer task relationships, whereas Human Knowledge and GPT-o1 rely solely on task descriptions, as the example illustrated in Fig.~\ref{fig:gas_filled_ballons}. This difference arises due to practical and computational constraints. First, for the human expert, manually analyzing thousands of student responses across tasks is infeasible due to the overwhelming scale and effort required \cite{koedinger2015data}. Similarly, GPT-o1, while capable of processing text-based task descriptions, lacks the capacity to analyze large volumes of student response data. Its context window constraints prevent it from incorporating full response data \cite{jaech2024openai}, making it unable to derive insights from response patterns.
We further note that GW-SMM relies only on student responses because incorporating question text could introduce textual bias \cite{gururangan2018annotation}, making the model overly dependent on question wording and limiting its ability to classify responses independently. Moreover, all of our task questions contain images or tables that the text-based BERT model cannot process, leading to incomplete representations \cite{devlin2018bert}. Given these constraints, we focus the GW-SMM on student responses to ensure unbiased learning from fully accessible data.

% To analyze the impact of model merging on accuracy, we conducted paired t-tests comparing the performance of different merging methods. The merging methods include GW-SMM (our proposed Gromov-Wasserstein Scoring Model Merging-based method), Human Knowledge (expert-driven merging), and GPT-o1 (prompting GPT-o1 to identify model similarities).
% Additionally, a repeated measures ANOVA was performed to evaluate the overall effect of merging across methods.

\begin{figure}[h]
    \centering
    \includegraphics[width=1\linewidth]{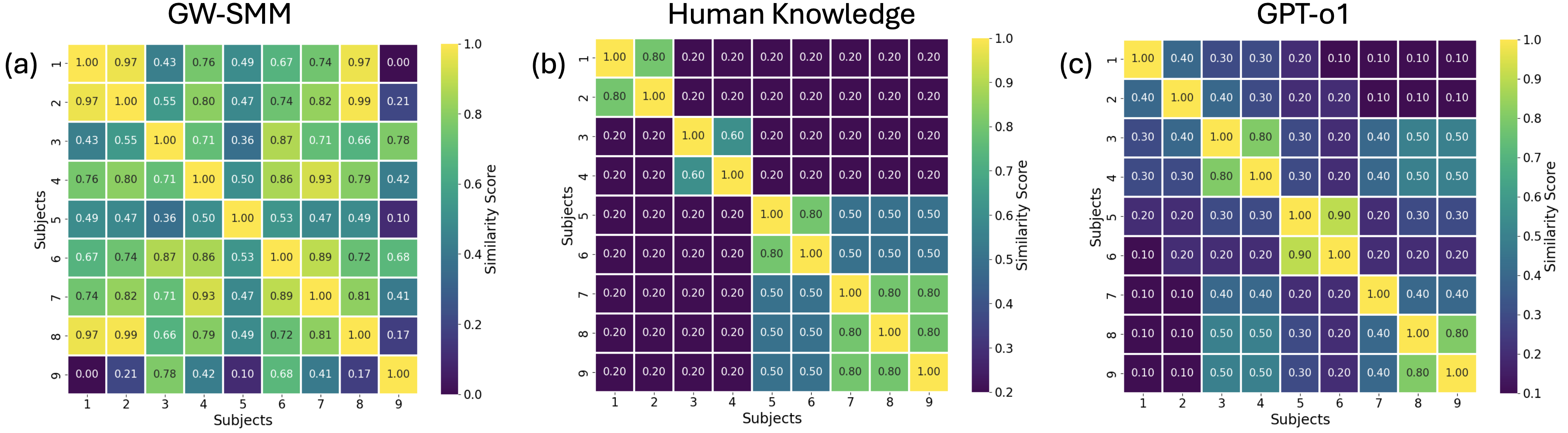}
    \caption{Heatmaps of the similarity scores of each method.}
    \label{fig:heatmap}
    \vspace{-6pt}
\end{figure}

% \begin{table}[h]
% \centering
% \caption{Merging plans of each method} 
% \label{tab:plans}
% % \vspace{5pt}
% \begin{tabular}{l|ccc|}
% \toprule
% \textbf{Method} & \textbf{Merge Plan} \\
% \midrule
% GW-SMM  & (1,2,8), (3), (4,6,7), (5), (9) \\
% GPT-o1 & (1), (2), (3,4,8,9), (5,6), (7) \\
% Human Knowledge & (1,2), (3), (4), (5,6), (7,8,9) \\
% % Combined GPT-o1 and GW-SMM & \{1,2\}, \{3,4\}, \{5,6\}, \{7,8\}, \{9\} \\
% % Combined Human Knowledge and GW-SMM & \{1,2\}, \{3,4,8\}, \{5,6\}, \{7\}, \{9\} \\
% \bottomrule
% \end{tabular}
% \vspace{5pt}
% \end{table}

%======================
\begin{figure}[htp!]
    \centering
    % % First row with three images
    % \begin{subfigure}{0.32\textwidth}
    %     \centering
    %     \includegraphics[width=\linewidth]{figs/plot_GW-SMM.png}
    %     \caption{GW-SMM (Ours)}
    % \end{subfigure}
    % \begin{subfigure}{0.32\textwidth}
    %     \centering
    %     \includegraphics[width=\linewidth]{figs/plot_Human Knowledge.png}
    %     \caption{Human Knowledge}
    % \end{subfigure}
    % \begin{subfigure}{0.32\textwidth}
    %     \centering
    %     \includegraphics[width=\linewidth]{figs/plot_ GPT-o1.png}
    %     \caption{Prompting GPT-o1}
    % \end{subfigure}

    % Second row with two images
    \begin{subfigure}{0.29\textwidth}
        \centering
        \includegraphics[width=\linewidth]{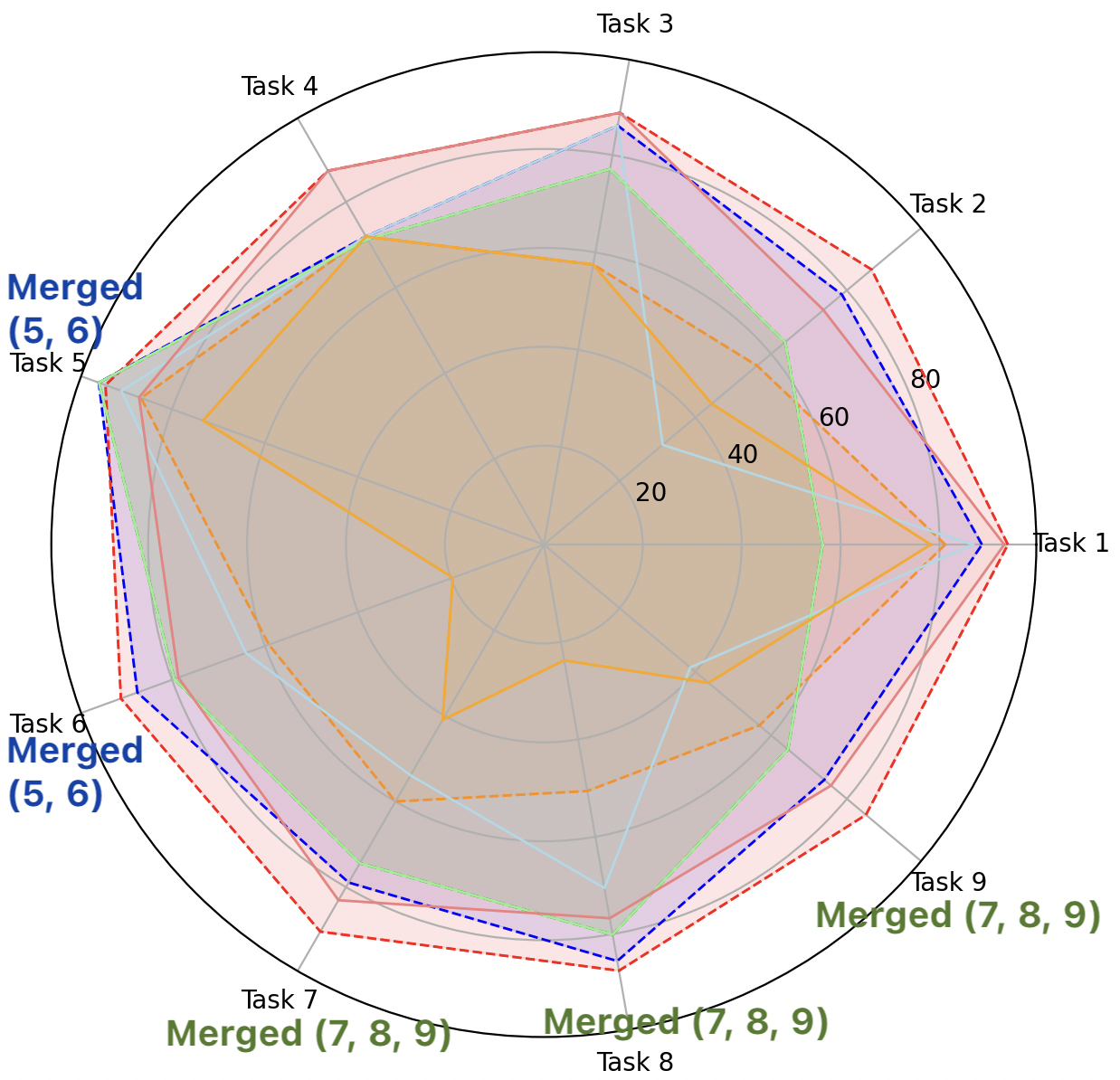}
        \caption{Human}
    \end{subfigure}
    \begin{subfigure}{0.29\textwidth}
        \centering
        \includegraphics[width=\linewidth]{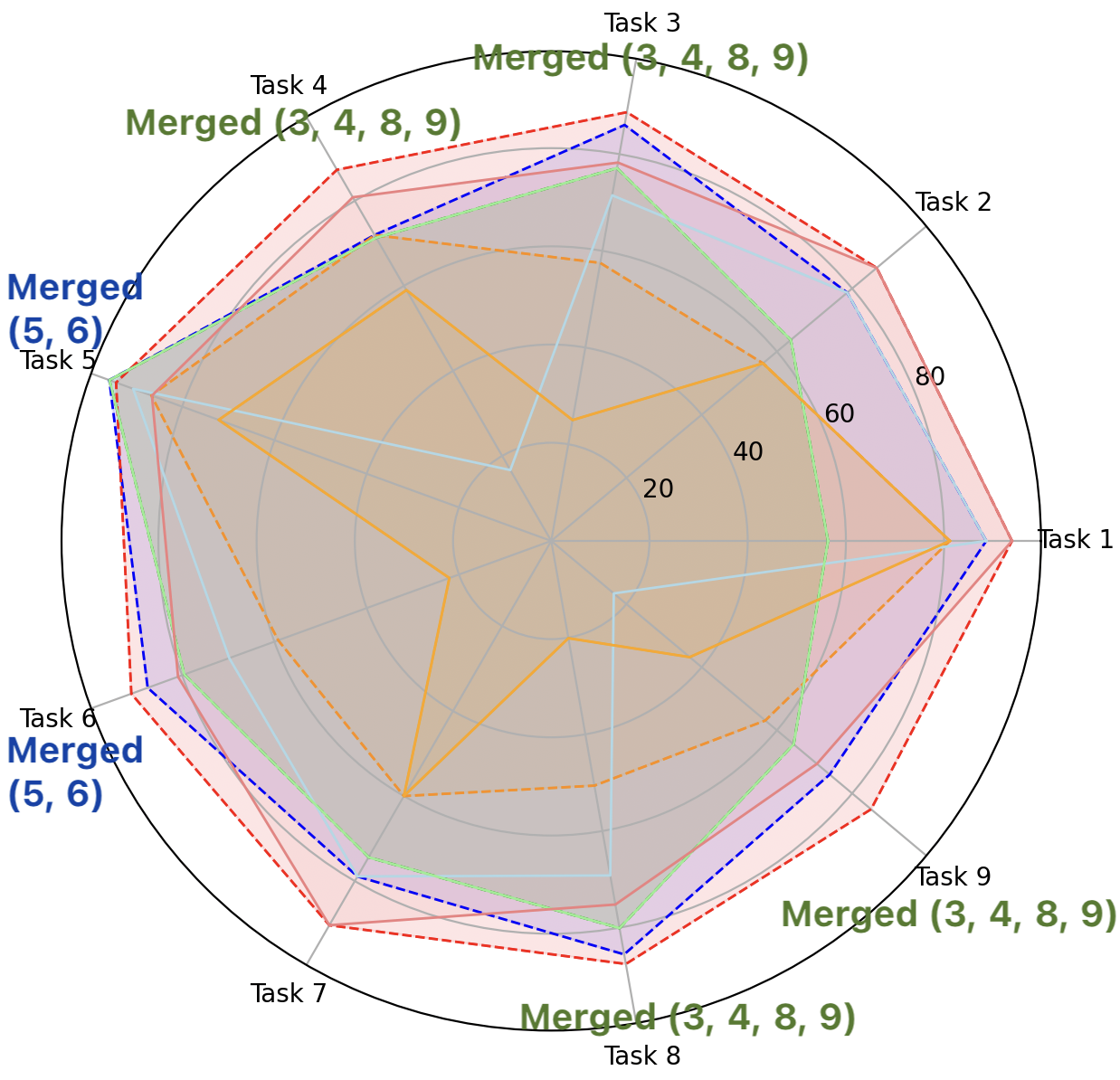}
        \caption{GPT-o1}
    \end{subfigure}
    \begin{subfigure}{0.40\textwidth}
        \centering
        \includegraphics[width=\linewidth]{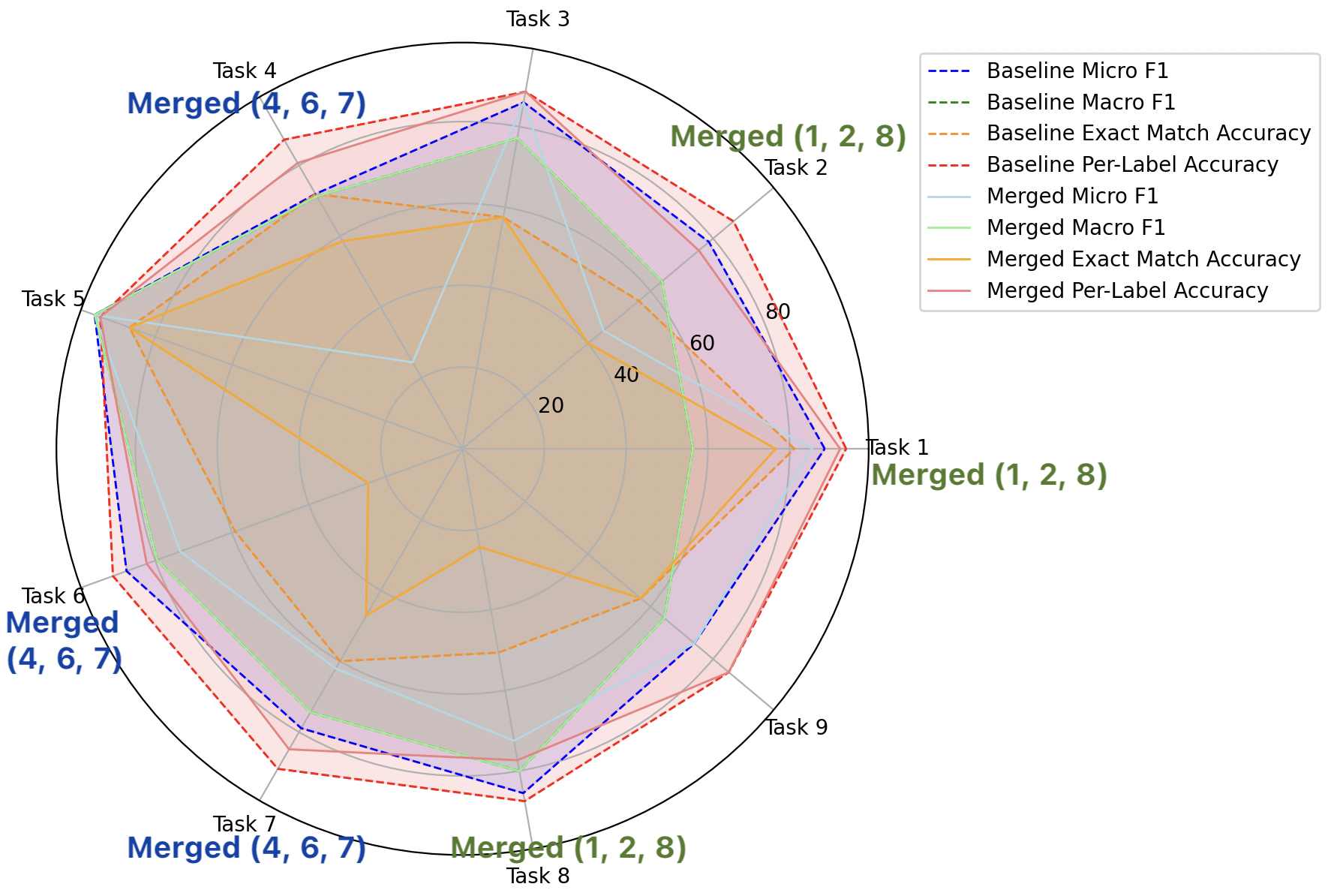}
        \caption{GW-SMM}
    \end{subfigure}

    \caption{Performance results before and after merging using three methods (GW-SMM (Ours), human knowledge, and GPT-o1) based merging.}
    \label{fig:combined_results}
\end{figure}
%======================

% \begin{table}[h]
%     \centering
%     \caption{Average performance comparison across evaluation metrics, with weighted means based on sample size. The best result for each metric is bolded among merging methods}
%     \label{tab:avg_comparison}
%     % \vspace{5pt}
%     \resizebox{0.8\textwidth}{!}{ 
%     \begin{tabular}{l|ccc|c}
%         \toprule
%         \textbf{Metrics} & \multicolumn{3}{c|}{\textbf{After-Merge}} & \textbf{Pre-Merge} \\
%         \cmidrule(lr){2-4}
%          & \textbf{GW-SMM } & \textbf{Human } & \textbf{GPT-o1 } &  \\
%         \midrule
%         Micro F1 Score       & \textbf{0.6872}  & 0.6741  & 0.6271  & 0.8291 \\
%         Macro F1 Score       & \textbf{0.5414}  & 0.5298  & 0.4895  & 0.7370 \\
%         Exact Match Accuracy & \textbf{0.5419}  & 0.5211  & 0.4883  & 0.6549 \\
%         Per-label Accuracy   & \textbf{0.8507}  & 0.8331  & 0.8255  & 0.8951 \\
%         \bottomrule
%     \end{tabular}
%     }
%     \vspace{-20pt}
% \end{table}

\begin{table}[h]
\centering
\caption{Merging plans and average performance comparison of each method. The best result for each metric is bolded among merging methods.}
\label{tab:merged_plans_performance}
\resizebox{\textwidth}{!}{ 
\begin{tabular}{l|c|cccc}
\toprule
\textbf{Method} & \textbf{Merge Plan} & \multicolumn{4}{c}{\textbf{After-Merge Performance}} \\
\cmidrule(lr){3-6}
 &  & \textbf{Micro F1} & \textbf{Macro F1} & \textbf{Exact Match} & \textbf{Per-label accuracy} \\
\midrule
GW-SMM  & (1,2,8), (4,6,7)  & \textbf{0.6872} & \textbf{0.5414} & \textbf{0.5419} & \textbf{0.8507} \\
GPT-o1  & (3,4,8,9), (5,6)  & 0.6271 & 0.4895 & 0.4883 & 0.8255 \\
Human Knowledge  & (1,2), (5,6), (7,8,9)  & 0.6741 & 0.5298 & 0.5211 & 0.8331 \\
\midrule
Pre-Merged & - & 0.8291 & 0.7370 & 0.6549 & 0.8951\\
\bottomrule
\end{tabular}
}
\end{table}

Fig. \ref{fig:heatmap} presents the heatmaps of similarity scores computed by GW-SMM, Human, and GPT-o1.
Table \ref{tab:merged_plans_performance} presents the merging plans decided by each method based on the similarity matrices using the approach described in Section \ref{sec:mergingplan}, along with the average performance of each method across the four evaluation metrics. We report the weighted mean, where weights are proportional to the testing sample size to mitigate the impact of sample size imbalance. 
Results show that GW-SMM consistently achieves the best performance among merging methods, demonstrating its effectiveness in aggregating multiple sources while preserving predictive accuracy. Although merging leads to some performance degradation compared to the Pre-Merge baseline, GW-SMM mitigates this loss more effectively than Human and GPT-o1. Specifically, GW-SMM outperforms Human and GPT-o1 in Micro F1 Score (0.6872 vs. 0.6741 and 0.6271), Macro F1 Score (0.5414 vs. 0.5298 and 0.4895), Exact Match Accuracy (0.5419 vs. 0.5211 and 0.4883), and Per-label Accuracy (0.8507 vs. 0.8331 and 0.8255), making it the most reliable merging strategy. While Pre-Merge results remain the upper bound, GW-SMM provides the best trade-off between merging efficiency and accuracy preservation.
% The smaller performance drop compared to alternative methods highlights GW-SMM’s superiority in balancing informativeness and consistency in merged outputs.

\begin{table}[h]
    \centering
    \caption{Statistical comparison of methods across metrics. Each value shows the t-statistic and p-value for pairwise comparisons among GW-SMM, Human Knowledge, and GPT-o1. Statistically significant results ($p < 0.05$) are bolded.}
    \resizebox{\textwidth}{!}{ 
    \begin{tabular}{l|cc|cc|cc}
        \toprule
        \textbf{Metrics} & 
        \multicolumn{2}{c|}{\textbf{GW-SMM vs Human}} & 
        \multicolumn{2}{c|}{\textbf{GW-SMM vs GPT-o1}} & 
        \multicolumn{2}{c}{\textbf{Human vs GPT-o1}} \\
        \cmidrule(lr){2-3} \cmidrule(lr){4-5} \cmidrule(lr){6-7}
        & \textbf{T-Stat} & \textbf{P-Value} & \textbf{T-Stat} & \textbf{P-Value} & \textbf{T-Stat} & \textbf{P-Value} \\
        \midrule
        Micro F1 Score       & 0.869983  & 0.391454  & \textbf{2.106563}  & \textbf{0.043925}  & 1.190602  & 0.243465 \\
        Macro F1 Score       & -0.187336 & 0.852702  & 1.273012  & 0.213125  & 1.604841  & 0.119365 \\
        Exact Match Accuracy & 1.363881  & 0.183098  & 1.173100  & 0.250302  & 0.062403  & 0.950670 \\
        Per-label Accuracy   & 1.577693  & 0.125483  & \textbf{2.637638}  & \textbf{0.013281}  & 0.769699  & 0.447703 \\
        \bottomrule
    \end{tabular}
    }
    \label{tab:stat_comparison}
\end{table}

Table~\ref{tab:stat_comparison} presents paired t-test results comparing the three methods, accounting for task-specific variations. The statistical analysis evaluates GW-SMM, Human Knowledge, and GPT-o1 across four evaluation metrics. 
The pairwise comparisons reveal that GW-SMM and GPT-o1 exhibit statistically significant differences on two metrics: Micro F1 Score ($t = 2.107$, $p = 0.044$) and Per-label Accuracy ($t = 2.638$, $p = 0.013$). These results indicate that GW-SMM outperforms GPT-o1 in these aspects. In contrast, no statistically significant differences ($p > 0.05$) were observed between GW-SMM and Human Knowledge or between Human Knowledge and GPT-o1 across any metric. For instance, comparisons between GW-SMM and Human Knowledge yielded $p$-values ranging from 0.126 to 0.853, while Human Knowledge versus GPT-o1 resulted in $p$-values between 0.120 and 0.951. The combined performance of all three methods is illustrated in Fig.~\ref{fig:combined_results}.
These results indicate that GW-SMM achieves superior performance compared to GPT-o1, with significant improvements in Micro F1 Score and Per-label Accuracy. While its performance exceeds that of Human Knowledge, the difference is not statistically significant. This suggests that GW-SMM effectively leverages model merging to achieve robust and efficient scoring, performing comparable with or even surpassing Human Knowledge in certain cases.

Each merging method combined different sets of models, reducing model storage and deployment costs. For instance, GW-SMM merged models (1,2,8), reducing storage overhead by a factor of three. Given that each original model was based on BERT with an approximate size of 450MB, this results in a 3x reduction in storage requirements. Similarly, other methods achieved comparable reductions, demonstrating that model merging offers significant efficiency improvements without substantial losses in accuracy.

\section{Conslusion}

This paper introduces GW-SMM, a novel approach to model merging that leverages the GW distance to align feature distributions across multiple automated scoring models. Our experimental results demonstrate that GW-SMM outperforms both human expert knowledge and GPT-o1-based merging approaches across multiple evaluation metrics, with statistically significant improvements in micro F1 score and per-label accuracy compared to GPT-o1-based merging. This method successfully reduced storage requirements by up to a factor of three while maintaining scoring performance.

A key advantage of our approach is its data-driven nature, which enables it to capture complex relationships between tasks that may not be apparent from surface-level analysis of task descriptions alone. While computational scalability remains a challenge as the number of tasks increases, GW-SMM represents a significant step forward in making automated scoring systems more practical and efficient to deploy at scale, while maintaining the high level of accuracy needed for educational applications.

While our merged models demonstrate strong performance, we observe there is still a notable gap compared to the pre-merge baseline, particularly in the micro F1 score (0.6872 vs. 0.8291) and macro F1 score (0.5414 vs. 0.7370). In future work, we aim to narrow this gap by incorporating advanced model merging algorithms that can better preserve task-specific features while maintaining shared knowledge across related tasks, such as applying adaptive layer-wise fusion strategies \cite{yang2023adamerging}.

% \subsubsection{Data and Ethics Statement.} 
% \textit{Data collection, reporting, and analysis.}  Due to the geographical diversity of participating teachers, the dataset is
% considered representative of the broader US middle school student population.
% \textit{Ethics.}

\subsubsection{\discintname}
The authors have no competing interests to declare that are relevant to the content of this article.

\newpage
\bibliographystyle{splncs04}

\bibliography{ref}
%

% Author, F.: Article title. Journal \textbf{2}(5), 99--110 (2o16)

% F., Editor, S. (eds.) CONFERENCE 2o16, LNCS, vol. 9999, pp. 1--13.
% Springer, Heidelberg (2o16). \doi{10.10007/1234567890}


% on Proceedings, pp. 1--2. Publisher, Location (2o10)

% LNCS Homepage, \url{http://www.springer.com/lncs}, last accessed 2023/10/25
% \end{thebibliography}

\end{document}